\documentclass[10pt, conference, compsocconf]{IEEEtran}

\usepackage{amsmath,amssymb,amsfonts}
\usepackage{subfig}
\usepackage{graphicx}
\usepackage{algorithm}
\usepackage{multirow}
\usepackage{xcolor}
\usepackage[noend]{algpseudocode}
\usepackage{textcomp}
\usepackage{hyperref}
\hypersetup{
    colorlinks=true,
    linkcolor=blue,
    filecolor=magenta,      
    urlcolor=cyan,
}

\begin{document}

\title{CardioCaps: Attention-based Capsule Network \\ for Class-Imbalanced Echocardiogram Classification}

\author{\IEEEauthorblockN{Hyunkyung Han\IEEEauthorrefmark{1},
Jihyeon Seong\IEEEauthorrefmark{1}, and
Jaesik Choi\IEEEauthorrefmark{2}}
\IEEEauthorblockA{\IEEEauthorrefmark{1}Korea Advanced Institute of Science and Technology (KAIST),
South Korea\\ 
}
\IEEEauthorblockA{\IEEEauthorrefmark{2}Korea Advanced Institute of Science and Technology (KAIST), INEEJI,
South Korea\\
Email: \{hkhan, jihyeon.seong, jaesik.choi\}@kaist.ac.kr}
}

\maketitle

\begin{abstract}
Capsule Neural Networks (CapsNets) is a novel architecture that utilizes vector-wise representations formed by multiple neurons. Specifically, the Dynamic Routing CapsNets (DR-CapsNets) employ an affine matrix and dynamic routing mechanism to train capsules and acquire translation-equivariance properties, enhancing its robustness compared to traditional Convolutional Neural Networks (CNNs). Echocardiograms, which capture moving images of the heart, present unique challenges for traditional image classification methods. In this paper, we explore the potential of DR-CapsNets and propose \textit{CardioCaps}, a novel attention-based DR-CapsNet architecture for class-imbalanced echocardiogram classification. CardioCaps comprises two key components: a weighted margin loss incorporating a regression auxiliary loss and an attention mechanism. First, the weighted margin loss prioritizes positive cases, supplemented by an auxiliary loss function based on the Ejection Fraction (EF) regression task, a crucial measure of cardiac function. This approach enhances the model's resilience in the face of class imbalance. Second, recognizing the quadratic complexity of dynamic routing leading to training inefficiencies, we adopt the attention mechanism as a more computationally efficient alternative. Our results demonstrate that CardioCaps surpasses traditional machine learning baseline methods, including Logistic Regression, Random Forest, and XGBoost with sampling methods and a class weight matrix. Furthermore, CardioCaps outperforms other deep learning baseline methods such as CNNs, ResNets, U-Nets, and ViTs, as well as advanced CapsNets methods such as EM-CapsNets and Efficient-CapsNets. Notably, our model demonstrates robustness to class imbalance, achieving high precision even in datasets with a substantial proportion of negative cases. The official code is available at \url{https://github.com/jihyeonseong/CardioCaps.git}. 
\end{abstract}


\IEEEpeerreviewmaketitle

\section{Introduction}

\begin{figure}
    \centering
    \includegraphics[clip, trim= 3.5cm 1cm 3cm 1cm, width=0.45\textwidth]{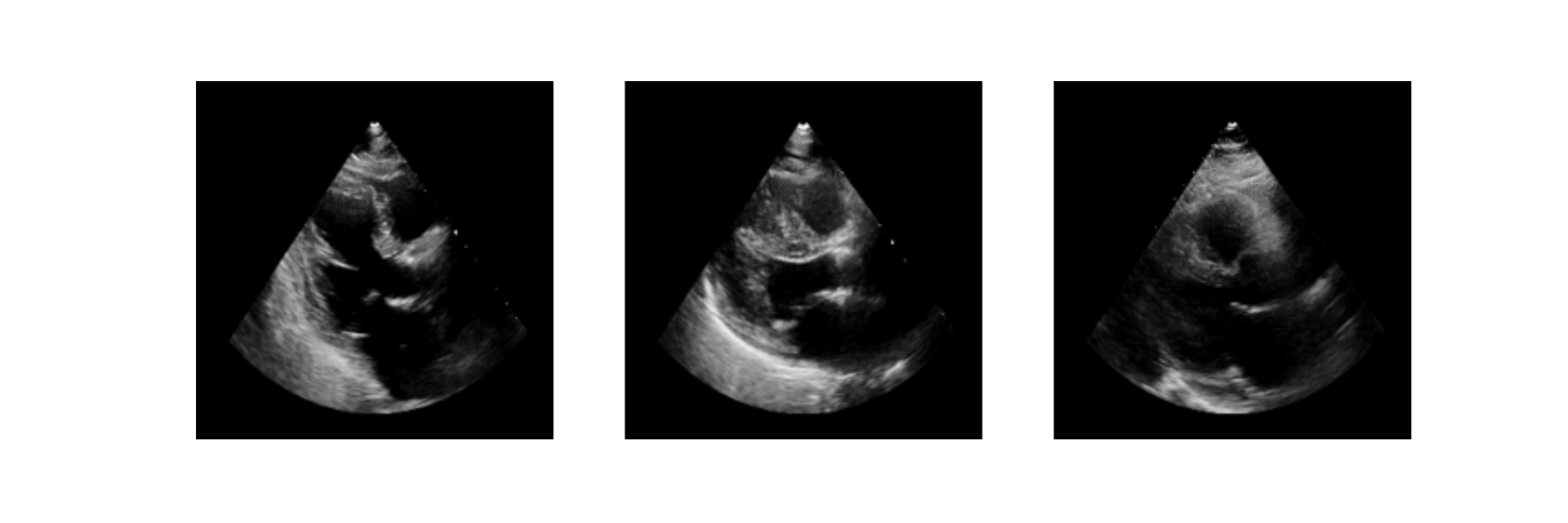}
    \caption{\textbf{Samples of echocardiogram datasets.} As shown in the figure, the heart is in motion, and its shape deviates slightly from the one confirmed by the doctor. Our study demonstrates that the DR-CapsNets, acting as a translation-equivariance learning model, can achieve optimal performance in diagnosing echocardiograms.}
    \label{fig:sample}
\end{figure}

\begin{figure*}
    \centering
    \includegraphics[width=0.85\textwidth]{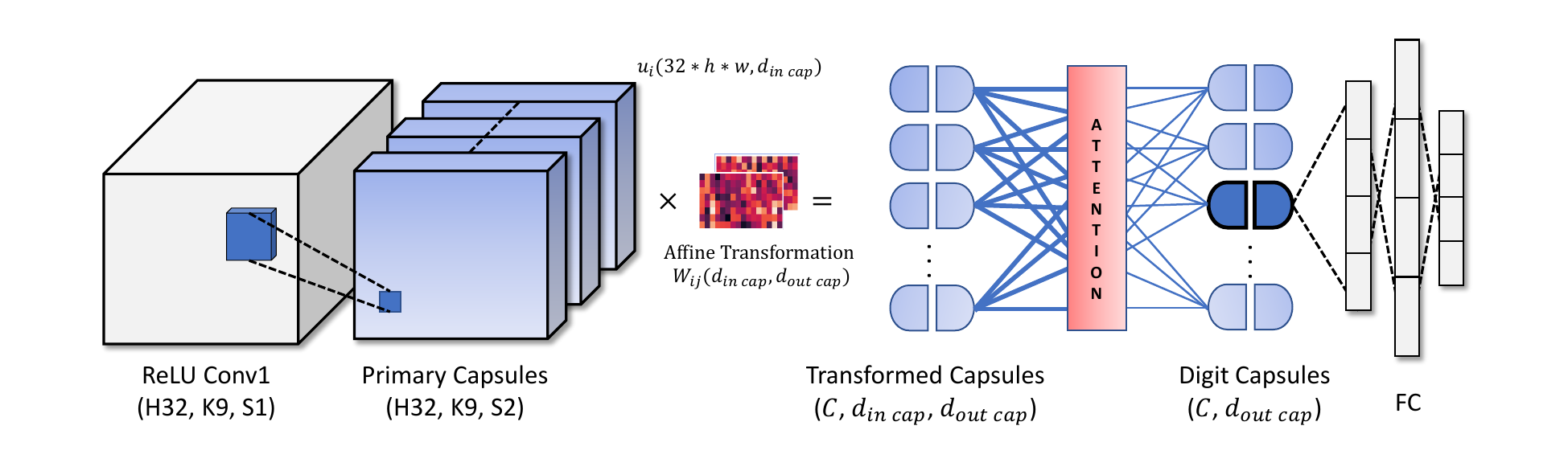}
    \caption{\textbf{CardioCaps: attention-based DR-CapsNets.} The diagram illustrates the architecture of the proposed attention-based DR-CapsNets, named CardioCaps, designed for class-imbalanced echocardiogram classification. The network comprises five components: ReLU Conv, Primary Capsules, Affine Transformation, Attention, and FC Layer Decoder. First, the ReLU Conv extracts features using a large kernel size. Subsequently, the Primary Capsule layer transforms feature neurons into capsules with a $d_{in cap}$ dimension. Third, the Affine Transformation matrix is applied to the capsules to ensure robust transformations. Fourth, we utilize the attention mechanism instead of dynamic routing for enhanced training efficiency. Finally, the normalized digit capsules are fed into the FC decoder to obtain reconstruction outputs. Note that $d$ represents the dimension of capsules, and $C$ denotes the unique class number.}
    \label{fig:main}
\end{figure*}

Accurate medical image classification plays a crucial role in disease diagnosis. Echocardiograms, capturing moving images of the heart, present unique challenges for traditional image classification methods \cite{lee2020point, baribeau2020handheld}. Their dynamic nature and complex spatial relationships often hinder the effectiveness of traditional Convolutional Neural Networks (CNNs). Capsule Neural Networks (CapsNets) is a promising method across diverse applications, emphasizing translation equivariance learning, particularly in dynamic medical imaging analysis like ultrasound \cite{sabour2017dynamic}. CapsNets introduces the concept of a capsule, a combination of neurons using vector-wise representation. These capsules encode essential information, including the object's angular orientation, spatial positioning, and scaling attributes. Dynamic Routing Capsule Networks (DR-CapsNets) employ an affine matrix and dynamic routing to train capsules, enabling the acquisition of translation-equivariance property. This property positions DR-CapsNets as a novel tool in numerous studies aiming to enhance the medical imaging analysis \cite{kavitha2021deep}.

However, DR-CapsNets exhibits two limitations: 1) it lacks robustness to imbalanced datasets, and 2) it involves the quadratic optimization complexity of dynamic routing. The margin loss in DR-CapsNets proves less robust when confronted with imbalanced datasets, a common scenario in the medical domain. This limitation stems from not considering the proportion of negative and positive classes within the loss function \cite{9506389}. To tackle this issue, we introduce a `weighted' margin loss along with an auxiliary loss function for the regression task based on Ejection Fraction (EF). The class imbalance ratio is incorporated as the weight to emphasize positive cases. In addition, we employ the Mean Squared Error (MSE) loss function for the downstream regression task, also known as the L2 loss function, with decay to regulate model training. Consequently, the weighted margin loss measures the disparity between predicted and actual labels, while the L2 regularization loss prevents overfitting to the dominant class, enhancing model robustness on imbalanced datasets.

Dynamic routing enables the learning of translation equivariance, standing out as one of the most critical components in DR-CapsNets. Dynamic routing focuses on learning hierarchical capsule relationships, unlike CNNs' pooling mechanism, which induces translation invariance learning. This hierarchical relationship is learned by calculating coupling coefficients representing the similarity between capsules. However, the quadratic optimization complexity  $\mathcal{O}(N^2)$ of dynamic routing poses a challenge to training efficiency since dynamic routing accumulatively calculates similarity between capsules in an iterative manner. To address this inefficiency, we employ the attention mechanism \cite{nips2017attention} to calculate the similarity between the upper and lower layers of capsules. This process closely resembles dynamic routing, involving the calculation of coupling coefficients through a dot product with softmax.

Additionally, we hypothesize that the affine matrix of DR-CapsNets is crucial for the robustness in different angles captured within the same image on an echocardiogram dataset. To test this hypothesis, we conduct an ablation study based on \cite{gu_cvpr}. Instead of using the original shared learnable parameters, we replace the affine matrix with a convolution layer and constant parameters fixed to 1. This study provides evidence supporting the robustness of DR-CapsNet against echocardiograms captured at different angles through the shared affine matrix.

The proposed attention-based DR CapsNets, named \textit{CardioCaps}, coupled with the novel loss function, is evaluated on an echocardiogram dataset \cite{ouyang2020video}. The dataset exhibits class imbalance, with 80\% normal echocardiograms and 20\% abnormal ones. CardioCaps achieves an impressive accuracy of 85\%, surpassing the original DR-CapsNets' accuracy of 57\%. CardioCaps outperforms state-of-the-art baselines in comparison with various models, including machine learning (Logistic Regression, Random Forest \cite{randomforest}, XGBoost \cite{xgboost}), deep learning (CNNs \cite{cnn}, ResNet \cite{resnet}, U-Net \cite{unet}, ViTs \cite{vit}), and advanced CapsNets methods (EM-CapsNets \cite{e2018matrix}, Efficient-CapsNets \cite{Mazzia_2021}). CardioCaps achieves high precision even in datasets with a substantial proportion of negative cases.

Therefore, our contributions are summarized as follows:
\begin{itemize}
\item We introduce DR-CapsNets to the challenging problem of echocardiogram diagnosis.
\item We propose a new loss function incorporating a weighted margin loss and L2 regularization loss to handle imbalanced classes in echocardiogram datasets.
\item We employ an attention mechanism instead of dynamic routing to achieve training efficiency.
\item We demonstrate the robustness of CardioCaps through comprehensive comparisons against various baselines.
\end{itemize}

\section{Background and Related Works}
\subsection{Echocardiogram}
\label{sec2.2}

\begin{algorithm}
    \caption{Dynamic Routing}
    \hspace*{\algorithmicindent} \textbf{Input1: } for all capsule $i$ in layer $l$: $\mathbf{C}_i$ \\
    \hspace*{\algorithmicindent} \textbf{Input2: } for all capsule $j$ in layer $(l+1)$: $\mathbf{C}_j$\\
    \hspace*{\algorithmicindent} \textbf{Output: $v_j$} 
    \begin{algorithmic}[1]
    \Procedure{Routing}{$\hat{u}_{j|i}, r, l$}
        
        for all $\mathbf{C}_{ij}$ : $b_{ij} \leftarrow 0.$ 
        
        \For{$r$ iterations}
    
            \State for all $\mathbf{C}_i$: $c_i \leftarrow$ softmax($b_i$)
    
            \State for all $\mathbf{C}_j$: $s_j \leftarrow \sum_i c_{ij} \hat{u}_{j|i}$
    
            \State for all $\mathbf{C}_j$: $v_j \leftarrow$ squash($s_j$) 
    
            \State for all $\mathbf{C}_{ij}$: $b_{ij} \leftarrow b_{ij} + \hat{u}_{j|i} \cdot v_j$
    
        \EndFor
        
        \Return $v_j$
    
    \EndProcedure
    \end{algorithmic}
    \label{algo}
\end{algorithm}

Echocardiogram is a non-invasive imaging technique that uses ultrasound to produce images of the heart. It is a widely used diagnostic tool for a variety of heart conditions, including heart failure, valvular heart disease, and congenital heart disease \cite{cahalan2002american}. It is employed to diagnose cardiomyopathies and valvular heart diseases \cite{ratnayaka2013real}. Recent advancements in deep learning have led to its application in echocardiogram image analysis, showing promising results. Deep learning models can be trained to understand the complex relationships between different structures in the heart and their corresponding pathologies, ultimately improving the accuracy and efficiency of echocardiogram diagnosis.

However, inherent variability in images poses challenges in developing deep learning models for echocardiogram \cite{scholl2011challenges}. Echocardiogram images are acquired from various angles, including apical view, parasternal view, and subcostal view, and exhibit significant variations in the appearance of the heart. Each of these views provides a different perspective of the heart, contributing to what is known as various angle proving. Therefore, it is crucial for deep learning models to learn to identify the heart and its structures from all of these different angles.

Another significant challenge is the class imbalance problem encountered in most echocardiogram datasets. These datasets contain a considerably higher number of normal images compared to abnormal ones. The deep learning model encounters difficulties when accurately identifying abnormal images, especially in datasets with substantial imbalances. Therefore, addressing the class imbalance problem becomes crucial to ensure that deep learning models can effectively learn and precisely identify abnormal images.

There are two strategies to tackle the class imbalance problem in model learning. First, oversampling abnormal images or undersampling normal images in the training dataset generates a balanced distribution by creating copies of underrepresented images. Second, a weighted loss function can be employed, assigning greater importance to abnormal images during training. This prioritizes learning from the minority class. However, both strategies carry a potential risk of overfitting \cite{razzak2018deep}. 

\subsection{Dynamic Routing Capsule Neural Network}
Convolutional Neural Networks (CNNs) \cite{cnn} stand out as a powerful method in computer vision, but they come with certain limitations. First, CNNs are designed to learn translation invariance, which may reduce their robustness to changes in the position or angle of features within an image. Second, pooling layers in CNNs decreases the spatial size of feature maps. The pooling method results in the loss of position-equivariance information, while it effectively reduces parameters and prevents overfitting in general classification tasks. In short, CNNs face challenges in accurately classifying objects at different positions and angles within an image.

To overcome these limitations, DR-CapsNets employs two distinct approaches: vector-wise representation, known as capsules, and dynamic routing as a replacement for pooling. Capsules can encode various features within an input image, which proves challenging for a single neuron with a scalar representation. Also, through dynamic routing, DR-CapsNets captures hierarchical spatial relationships between capsules. This unique design endows DR-CapsNets with improved translation-equivariance learning compared to CNNs.

In the detailed process of DR-CapsNets, the input image undergoes ReLU convolution and is then transformed into capsules within the primary capsule layer. Each primary capsule is subject to a transformation by casting a vote using a transformation matrix $W_{ij} \in R^{D_{in}\times N\ast D_{out}}$, where $N$ is the number of output classes, and $D_{\text{out}}$ represents the dimensions of the output capsules. The vote is mathematically expressed as $\hat{u}_{j|i}=u_i W_{ij}$. The routing mechanism considers all votes, determining coupling coefficient weights $c_{ij}$  for each vote $\hat{u}_{j|i}$. This routing process involves iterative steps outlined in Algorithm \ref{algo}. DR-CapsNets incorporates a squashing function, serving as the vector-wise representation of the sigmoid function: 
$$v_j = squash(s_j) = \frac{||s_j||^2}{1+||s_j||^2} \frac{s_j}{||s_j||}.$$
These components enable DR-CapsNets to achieve robustness to images captured from various angles \cite{gu_cvpr}.

\section{Empirical Studies on DR-Capsule Network}
\label{sec3}

\subsection{Experimental Settings}

\subsubsection{Echocardiogram Datasets}
EchoNet-LVH dataset \cite{ouyang2020video} comprises 12,000 parasternal-long-axis echocardiogram videos obtained from patients at Stanford Hospital, including clinical measurements, physician reports, and echocardiography calculations. The accredited echocardiography laboratory provided imaging for patients with diverse cardiac conditions such as atrial fibrillation, coronary artery disease, cardiomyopathy, aortic stenosis, and amyloidosis. Each video underwent cropping and masking to enhance relevance by eliminating extraneous information.

\subsubsection{Pre-processing}
Given the characteristics of echocardiogram videos, we begin by converting the video data into image format. We perform image cropping to extract additional features using a dataset comprising 300 individual videos. We implement two center-crop operations to focus on the region of interest precisely since the triangular shape of echocardiograms can present challenges in isolating the heart valve. As a result, our input shape becomes (9, 256, 256), achieved by resizing the original echocardiogram videos from (768, 1024) to (256, 256) and applying two center-crop operations to extract supplementary features.

\subsubsection{Hyper-parameters}
In Figure \ref{fig:main}, CardioCaps is implemented with the following parameters: 32 hidden dimensions, kernel sizes of 9, primary capsule dimensions of 8, and digit capsule dimensions of 16. The model is optimized using the Adam optimizer \cite{kingma2017adam} with a learning rate of $1e-4$, a batch size of 8, and trained for 100 epochs. Early stopping is applied with a patience step of 5 to prevent overfitting. The seed number is arbitrarily fixed to 10.

\subsection{Robustness to Class Imbalance Datasets}
The loss function in the original DR-CapsNets combines two components: margin loss and reconstruction loss \cite{sabour2017dynamic}. Margin loss uses the length of the instantiation vector to ensure that, when a specific class is present in the image, its corresponding class digit capsule has a long instantiation vector. Based on the reconstruction from the digit capsule, the Mean Squared Error (MSE) loss function is incorporated alongside the margin loss.

The margin loss function is defined as follows:
\begin{equation}
\begin{split}
    L_k = & T_k \max(0, m^+ - ||v_k||)^2 \\
    & + \lambda (1 - T_k) \max(0, ||v_k|| - m^-)^2,
\end{split}
\end{equation}
where $T_k = 1$ if class $k$ is present, $m^+ = 0.9$, $m^- = 0.1$, and $\lambda=0.5$ by default. However, the original DR-CapsNets loss function cannot address the class imbalance problem in the margin loss, a common issue in medical datasets \cite{ho2019real}. Relying solely on the margin loss may lead to overfitting to the dominant label.

To address this issue, we introduce a novel loss function tailored for echocardiogram diagnosis: a weighted margin loss function accompanied by an auxiliary loss based on a downstream regression task related to the Ejection Fraction (EF). First, we incorporate the class imbalance ratio between patients and healthy cases into the original margin loss function. Next, as the regression task for the regularization term, we utilize the width of the echocardiogram, determined through metrics such as LVSD and LVID. Therefore, the final loss function is defined as follows:
\begin{equation}
L = L_k \times p_k + \lambda_{reg} \frac{1}{N} \sum_N ||\hat{y}_{reg} - y_{reg}||_2,
\end{equation}
where $p$ represents the ratio of class imbalance, $p_k = c_k / N$ signifies the proportion of class $k$ labels with $c_k$ instance sizes out of the total dataset size $N$, and $y_{reg}$ denotes the width of the echocardiogram. The weighted margin loss and regression auxiliary loss enable robust training of the DR-CapsNets in highly class-imbalanced echocardiogram scenarios.

\subsubsection{Empirical Result Analysis}
\label{sec3.2.1}
\begin{figure}
    \centering
    \subfloat[original DR]{\includegraphics[width=0.3\linewidth]{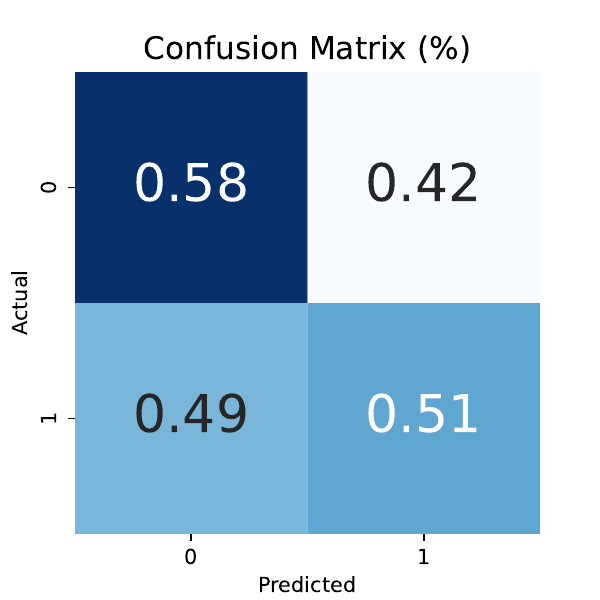}}
        \subfloat[w-margin]{\includegraphics[width=0.3\linewidth]{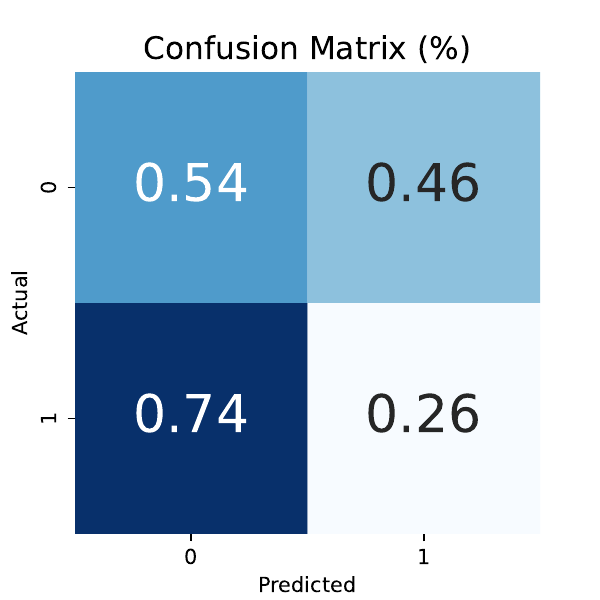}}
        \subfloat[w-margin\\ \& auxiliary loss (ours)]{\includegraphics[width=0.3\linewidth]{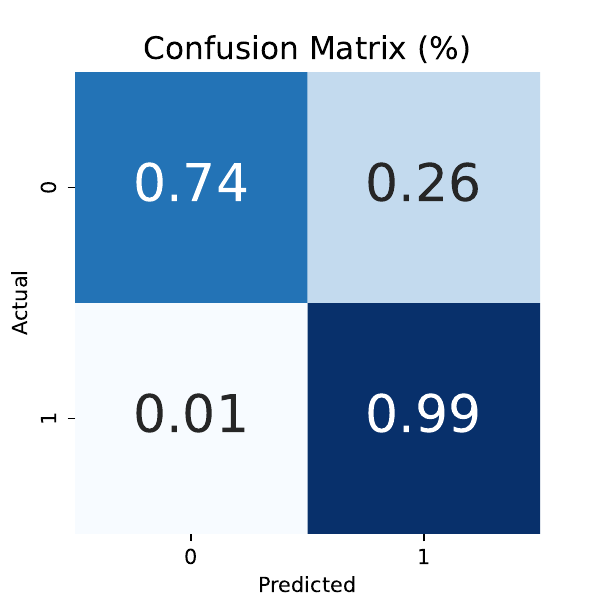}}
    \caption{Empirical study on loss function.}
    \label{fig:loss}
\end{figure}

To validate the importance of our new approach, we conduct an empirical study on the loss function. As presented in Table \ref{tab:loss}, our novel loss function demonstrates superior performance in achieving robustness on class-imbalanced echocardiogram datasets. In detail, as illustrated in Figure \ref{fig:loss}, the original DR-CapsNets with the margin loss function fall short in addressing class imbalance issues, resulting in a high false positive error rate. The weighted margin loss still falls short with a high false positive error rate, even when attempting to capture positive samples. On the other hand, our new loss function has effectively learned the positive class while maintaining a low false positive ratio, as evidenced by a high ROC score, which incorporates weighted margin loss and an auxiliary loss function from the EF regression task with an optimal $\lambda$ decay of 0.05. Nevertheless, a limitation persists due to a relatively high false negative ratio, resulting in lower precision with an absolute number of patients of 168, false negatives of 246, and true positives of 167.

\begin{table}[t]
    \centering
    \resizebox{\linewidth}{!}{
    \begin{tabular}[b]{c|ccc}
        \hline
        \multirow{2}{*}{Methods} & original DR & DR & DR weighted margin \\
        & -margin loss & weighted margin loss & \& auxiliary loss (ours) \\
        \hline
        Accuracy & 0.5707 & 0.4981 & \bf 0.7763\\
        F1 score & 0.6310 & 0.5687 & \bf 0.8066\\
        ROC AUC & 0.5441 & 0.3988 & \bf 0.8656\\
        PR AUC & 0.1786 & 0.0911 & \bf 0.4044\\
        \hline
    \end{tabular}}
    \caption{This table demonstrates that the new loss function, composed of weighted margin loss and an auxiliary loss from the EF regression task, can robustly train DR-CapsNets on an imbalanced echocardiogram dataset.}
    \label{tab:loss}
\end{table}

\subsection{Dynamic Routing and Attention}
Dynamic routing is a process that iteratively computes dot product similarities between capsules in the lower and upper layers, ultimately determining the digit capsule based on the accumulated coupling coefficients. However, this iterative calculation has an optimization complexity of $\mathcal{O}(N^2)$, which is highly inefficient. Additionally, according to \cite{gu_cvpr}, the iterative nature of dynamic routing may harm transformation robustness. To tackle this challenge, alternative and more efficient methods have been explored \cite{wang2018an, Mazzia_2021}. Among them, we specifically focus on the attention mechanism \cite{nips2017attention} as a potential replacement for dynamic routing.

Attention is a method that measures similarity by computing the dot product between the key $K$ and query $Q$ and then assigns scores through the softmax function:
$$Attention(Q, K, V) = softmax(\frac{QK^\top}{\sqrt{d_k}})V.$$
Multiplying the value $V$ by the calculated attention score weights the importance of different elements in the input. This is similar to the method for calculating the coupling coefficient of Equation \ref{cc}.

\begin{figure}
    \centering
        \subfloat[original DR]{\includegraphics[width=0.3\linewidth]{img/dr_original.pdf}}
        \subfloat[new loss (ours)]{\includegraphics[width=0.3\linewidth]{img/dr_weight_005.pdf}}
        \subfloat[new loss\\ \& attention (ours)]{\includegraphics[width=0.3\linewidth]{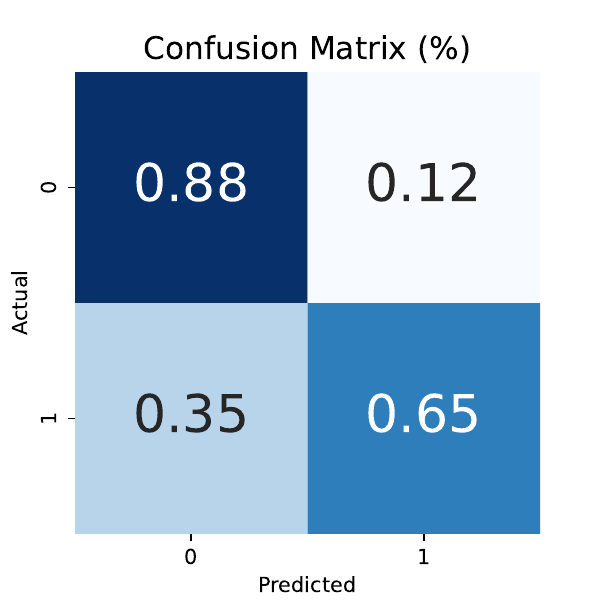}}
    \caption{Empirical study on dynamic routing and attention.}
    \label{fig:attention}
\end{figure}

\begin{table}[t]
    \centering
    \resizebox{\linewidth}{!}{
    \begin{tabular}[b]{c|ccc}
        \hline
        \multirow{2}{*}{Methods} & original DR & DR & DR new loss \\
        & -margin loss & new loss (ours) & \& attention (ours)  \\
        \hline
        Accuracy & 0.5707 & 0.7763 & \bf 0.8496\\
        F1 score & 0.6310 & 0.8066 & \bf 0.8573\\
        ROC AUC & 0.5441 & \bf 0.8656 & 0.7697\\
        PR AUC & 0.1786 & 0.4044 & \bf 0.5046\\
        \hline
    \end{tabular}}
    \caption{This table illustrates that the utilization of the attention mechanism, as opposed to dynamic routing, leads to more robust learning in class-imbalanced echocardiogram datasets, achieving higher precision.}
    \label{tab:attention}
\end{table}

\begin{equation}
    c_{ij}^{(t+1)} = \frac{exp(b_{ij} + \sum_{r=1}^t v_j^{(r)} \hat{u}_{j|i})}{\sum_k exp(b_{ik} + \sum_{r=1}^t v_k^{(r)} \hat{u}_{k|i})}.
    \label{cc}
\end{equation}

The distinction between the attention mechanism and dynamic routing lies in the accumulation of similarity. In dynamic routing, the coupling coefficient is computed by multiplying the accumulated score by $v$. However, we assume that this cumulative scoring might harm transformation robustness, as early mentioned by \cite{gu_cvpr}. Therefore, we replace dynamic routing with convolutional attention layer, $conv(\hat{u}_{j|i})$, and the final capsules are calculated as follows,

\begin{equation}
    v_j = squash( softmax( conv(\hat{u}_{j|i})) \cdot \hat{u}_{j|i}).
\end{equation}

This convolutional attention layer enables the DR-CapsNets to output the digit capsule in a non-iterative manner.

\subsubsection{Empirical Result Analysis}
We conduct comparison experiments to demonstrate the need to replace dynamic routing with attention. As presented in Table \ref{tab:attention}, DR-CapsNets with attention and new loss function achieve the highest accuracy, F1 score, and PR AUC, except for the ROC AUC score. However, low precision results in an increased false-negative error rate, as discussed in Section \ref{sec3.2.1}. This implies that alarms occur more frequently than the actual number of patients. As shown in Figure \ref{fig:attention}, we affirm that the attention method replacing dynamic routing achieves robust performance, exhibiting the highest precision score with an absolute number of false negatives of 108 and true positives of 110. In conclusion, DR-CapsNets with our new loss function and attention mechanism, named CardioCaps, attain the highest level of robustness in imbalanced echocardiogram datasets.

\begin{figure}
    \centering
        \subfloat[original]{\includegraphics[width=0.3\linewidth]{img/dr_weight_05_attention.pdf}}
        \subfloat[conv]{\includegraphics[width=0.3\linewidth]{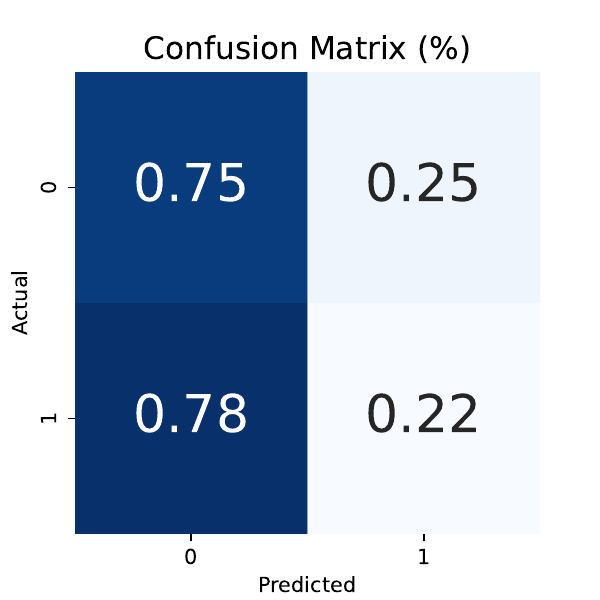}}
        \subfloat[constant]{\includegraphics[width=0.3\linewidth]{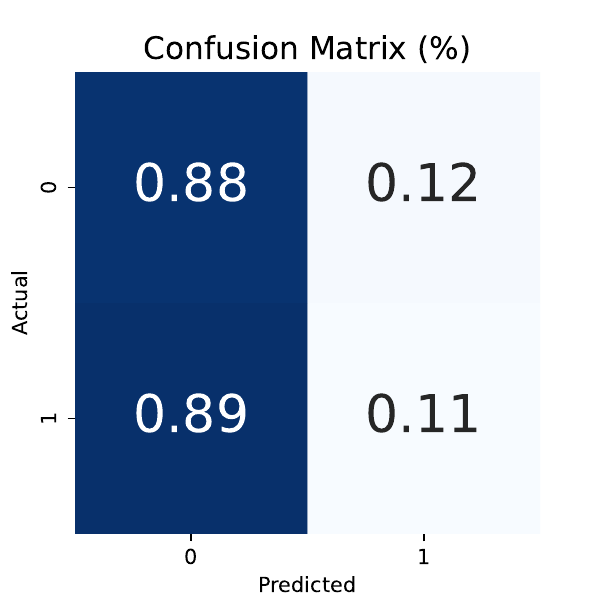}}
    \caption{Empirical study on affine matrix.}
    \label{fig:affine}
\end{figure}

\begin{table}[t]
    \centering
    \resizebox{\linewidth}{!}{
    \begin{tabular}[b]{c|ccc}
        \hline
        \multirow{2}{*}{Methods} & \multirow{2}{*}{original affine} & \multirow{2}{*}{conv affine} & \multirow{2}{*}{constant affine} \\
        &   \\
        \hline
        Accuracy & \bf 0.8496 & 0.6721 & 0.7654\\
        F1 score & \bf 0.8573 & 0.7005 & 0.7516\\
        ROC AUC & \bf 0.8573 & 0.4867 & 0.4953\\
        PR AUC & \bf 0.5046 & 0.1381 & 0.1417\\
        \hline
    \end{tabular}}
    \caption{This table demonstrates that the shared affine matrix of CardioCaps robustly facilitates translation equivariance learning.}
    \label{tab:affine}
\end{table}
    
\subsection{Translation Equivariance and Affine Matrix}

\subsubsection{Empirical Result Analysis}
To investigate the role of the affine matrix in achieving translation equivariance in CardioCaps, we hypothesize that the original shared affine matrix enables CardioCaps to learn this property, contributing to its state-of-the-art performance on echocardiogram datasets. To test this hypothesis, we conduct an ablation study, replacing the original shared affine matrix from DR-CapsNets with a convolutional layer and constant vectors, respectively. We then evaluate the performance of each model on the echocardiogram dataset under the same conditions with the new loss function and attention mechanism. As shown in Table \ref{tab:affine} and Figure \ref{fig:affine}, the results demonstrate that CardioCaps with a shared affine matrix outperform those with a convolutional layer and constant vectors. This highlights the crucial role of the original affine matrix from DR-CapsNets in achieving translation equivariance.

\begin{table*}[t]
    \centering
    \resizebox{\linewidth}{!}{
    \begin{tabular}{c|ccc|ccc|ccc||c}
         \hline
         \multirow{2}{*}{Methods} & \multicolumn{3}{c|}{Logistic Regression} & \multicolumn{3}{c|}{Random Forest \cite{randomforest}} & \multicolumn{3}{c||}{XGBoost \cite{xgboost}} & \multirow{2}{*}{CardioCaps} \\
         & Oversampling & Undersampling & Class Weight
         & Oversampling & Undersampling & Class Weight
         & Oversampling & Undersampling & Class Weight
         \\
         \hline
         Accuracy & $0.8292_{\pm 0.0149}$ & $0.7758_{\pm 0.0122}$ & $0.8292_{\pm 0.0186}$ & $0.8428_{\pm 0.0077}$ & $0.7768_{\pm 0.0023}$ & $0.8460_{\pm 0.0027}$ & $0.8365_{\pm 0.0005}$ & $0.7808_{\pm 0.0027}$ & $0.8502_{\pm 0.0025}$ & $\mathbf{0.8551}_{\pm 0.0054}$\\ 
         F1 score & $0.3982_{\pm 0.0574}$ & $0.5192_{\pm 0.0182}$ & $0.3873_{\pm 0.0637}$ & $0.4736_{\pm 0.0027}$ & $0.5530_{\pm 0.0034}$ & $0.2670_{\pm 0.0224}$ & $0.3606_{\pm 0.0188}$ & $0.5494_{\pm 0.0022}$ & $0.4560_{\pm 0.0124}$ & $\mathbf{0.8563}_{\pm 0.0011}$\\
         ROC AUC & $0.6417_{\pm 0.0320}$ & ${0.7836}_{\pm 0.0134}$ & $0.6344_{\pm 0.0342}$ & $0.6876_{\pm 0.0028}$ & $\mathbf{0.8304}_{\pm 0.0025}$ & $0.5746_{\pm 0.0090}$ & $0.6178_{\pm 0.0101}$ & $0.8206_{\pm 0.0003}$ & $0.6710_{\pm 0.0096}$ & $0.7326_{\pm 0.0371}$\\
         PR AUC & $0.4286_{\pm 0.0580}$ & $0.3857_{\pm 0.0167}$ & $0.4273_{\pm 0.0727}$ & $0.4852_{\pm 0.0250}$ & $0.3976_{\pm 0.0029}$ & $0.4832_{\pm 0.0242}$ & $0.4450_{\pm 0.0057}$ & $0.4008_{\pm 0.0120}$ & $0.5100_{\pm 0.0100}$ & $\mathbf{0.5273}_{\pm 0.0227}$\\
         \hline
    \end{tabular}}
    \caption{This table demonstrates that CardioCaps outperforms other machine-learning baselines, employing oversampling, undersampling, and a class weight matrix to address the class imbalance issue in the echocardiogram dataset.}
    \label{tab:final_results_ml}
\end{table*}

\section{Experiments}
\subsection{Experimental Settings} 
We compare our proposed CardioCaps with three categories of baseline models: 1) machine learning models, including Logistic Regression, Random Forest, and XGBoost; 2) deep learning models, including CNNs, Residual Network (ResNet), U-Net, and Vision Transformer (ViT); and 3) advanced CapsNets-based models, including EM-CapsNets and Efficient CapsNets.

\subsubsection{Machine-Learning Baselines}
\begin{itemize}
    \item \verb|Logistic Regression| models the probability using the logistic function to transform input features into a value between 0 and 1. 
    \item \verb|Random Forest| \cite{randomforest} is an ensemble learning method that builds multiple decision trees during training and outputs their collective predictions.
    \item \verb|XGBoost| \cite{xgboost} implements gradient boosting, a technique that incrementally builds a predictive model by adding weak learners like decision trees in stages to correct the errors of the existing model.
\end{itemize}

\subsubsection{Deep-Learning Baselines}
\begin{itemize}
    \item \verb|CNN-(1)| (Convolutional Neural Networks) \cite{cnn} is a widely used architecture for image classification tasks, featuring multiple convolutional and pooling layers to extract features from input images. CNN-(1) uses pooling layers for each convolutional layer.
    \item \verb|CNN-(2)| features a single pooling layer positioned in the last convolutional layer, resembling the location of dynamic routing in DR-CapsNets.
    \item \verb|ResNet18| \cite{resnet} is built on the residual block, incorporating two convolutional layers with a shortcut connection to mitigate vanishing gradient issues and enable the training of deeper networks.
    \item \verb|U-Net| \cite{unet}, a dedicated CNN for medical image segmentation, is adapted for a binary classification task by modifying the decoder.
    \item \verb|ViT| \cite{vit} redefines image classification, treating images as a sequence of patches and using transformer encoders for scalability, robustness, and ease of training. The input image is divided into patches, and a transformer encoder learns global representations through self-attention and positional encoding.
\end{itemize}

\subsubsection{Advanced CapsNets Baselines}
\begin{itemize}
    \item \verb|EM-CapsNets| \cite{e2018matrix} utilizes a 4$\times$4 pose matrix as a capsule to encode the spatial relationship of entities and employs EM routing instead of dynamic routing to learn the part-whole relationship between entities.
    \item \verb|Efficient CapsNets| \cite{Mazzia_2021} improves the efficiency of capsule networks through a self-attention routing algorithm. Unlike iterative routing algorithms, self-attention mechanisms enable capsules to directly calculate compatibility with parent capsules in a single step.
\end{itemize}

These baseline models are commonly utilized in computer vision and have been employed in various medical imaging tasks. They offer a robust foundation for comparisons with more complex and advanced models.

\subsubsection{Hyper-parameters}
We conduct experiments under the same conditions by setting the deep network models' hyperparameters as follows: hidden dimension 32, Adam optimizer, learning rate of $1e-4$, batch size of 8, and early stopping with a patience step of 5. A seed test is also performed to demonstrate the robustness of batch shuffling. Note that we apply loss weight to all models to address the class imbalance issue.

\begin{table*}[t]
    \centering
    \begin{tabular}{c|ccccc||c}
         \hline
         Methods & CNN (1) \cite{cnn} & CNN (2) \cite{cnn} & ResNet18 \cite{resnet} & UNet \cite{unet} & ViT \cite{vit} & CardioCaps \\
         \hline
         Accuracy & $0.7736_{\pm 0.0091}$ & $0.7745_{\pm 0.0018}$ & $0.8130_{\pm 0.0023}$ & $0.8379_{\pm 0.0100}$ & $0.7631_{\pm 0.0041}$ & $\mathbf{0.8551}_{\pm 0.0054}$\\ 
         F1 score & $0.8031_{\pm 0.0077}$ & $0.8050_{\pm 0.0016}$ & $0.8336_{\pm 0.0021}$ & $0.8114_{\pm 0.0334}$ & $0.7915_{\pm 0.0043}$ & $\mathbf{0.8563}_{\pm 0.0011}$\\
         ROC AUC & $0.8249_{\pm 0.0127}$ & $\mathbf{0.8609}_{\pm 0.0047}$ & $0.8201_{\pm 0.0050}$ & $0.6577_{\pm 0.1577}$ & $0.7553_{\pm 0.0195}$ & $0.7326_{\pm 0.0371}$\\
         PR AUC & $0.3934_{\pm 0.0119}$ & $0.4017_{\pm 0.0027}$ & $0.4394_{\pm 0.0040}$ & $0.2310_{\pm 0.2310}$ & $0.3636_{\pm 0.0103}$ & $\mathbf{0.5273}_{\pm 0.0227}$\\
         \hline
    \end{tabular}
    \caption{This table demonstrates that CardioCaps outperforms other deep-learning baselines on the echocardiogram dataset, which exhibits class imbalance issues.}
    \label{tab:final_results}
\end{table*}

\subsection{Result Analysis}
\subsubsection{Comparison with ML-baselines}
As mentioned in Section \ref{sec2.2}, sampling and the class weight matrix are popular approaches to addressing the class imbalance problem. We explore these methods in machine-learning-based models and compare their performance with that of CardioCaps. As shown in Table \ref{tab:final_results_ml}, our findings demonstrate that CardioCaps outperforms most baseline models across all evaluation metrics, except when compared to Random Forest using the undersampling method. However, it exhibits lower precision than CardioCaps, indicating a predominant classification with the negative class. Therefore, we can conclude that CardioCaps effectively addresses the class-imbalance issue in echocardiogram datasets through the incorporation of a new loss function and attention mechanism.

\subsubsection{Comparison with DL-baselines}
To demonstrate the superior performance of CardioCaps, we conduct comparative experiments with several deep-learning baseline models. First, CNN-(1) and CNN-(2) exhibit differences in pooling layers, showing different performances. In the case of CNN-(1), aiming for translation invariance learning, there is pooling for each convolutional layer. However, for CNN-(2), pooling is used at the end of the convolutional stack to preserve equivariant features from convolution. As a result, as shown in Table \ref{tab:final_results}, we observe higher ROC AUC and PR AUC scores for CNN-(2). Next, when comparing CNN-(2) with ResNet, we notice that increasing the model's capacity leads to more robust performance. However, it still falls short in terms of the low precision score, resulting in a high number of false alarms exceeding the actual number of patients.

U-Net, originally designed for segmentation tasks, does not exhibit excellent performance in classification, and its performance variance is significant. This sensitivity becomes evident in the model's response to batch shuffling, indicating a limitation of U-Net. ViT is one of the most powerful methods in image classification tasks and represents the most state-of-the-art approach. However, the patching approach of ViT severely lacks the translation equivariance property. Therefore, ViT is unsuitable for the echocardiogram dataset captured from different angles. In summary, we demonstrate that the translation equivariance learning of CardioCaps is the most suitable and exhibits robust performance for addressing class imbalance in echocardiogram classification.

\begin{table}[t]
    \centering
    \resizebox{\linewidth}{!}{
    \begin{tabular}{c|cc||c}
         \hline
         Methods & EM-CapsNets \cite{e2018matrix} & Efficient CapsNets \cite{Mazzia_2021} & CardioCaps \\
         \hline
         Accuracy & $0.8478_{\pm 0.0000}$ & $0.5316_{\pm 0.0054}$ & $\mathbf{0.8551}_{\pm 0.0054}$\\ 
         F1 score & $0.7780_{\pm 0.0000}$ & $0.5934_{\pm 0.0496}$ & $\mathbf{0.8563}_{\pm 0.0011}$\\
         ROC AUC & $0.5000_{\pm 0.0000}$ & $0.5284_{\pm 0.0143}$ & $\mathbf{0.7326}_{\pm 0.0371}$\\
         PR AUC & $0.0000_{\pm 0.0000}$ & $0.1666_{\pm 0.0050}$ & $\mathbf{0.5273}_{\pm 0.0227}$\\
         \hline
    \end{tabular}}
    \caption{This table demonstrates that CardioCaps outperforms other advanced CapsNets-based models on the echocardiogram dataset, which exhibits class imbalance issues.}
    \label{tab:final_results_caps}
\end{table}

\begin{figure}
    \centering
    \includegraphics[clip, trim= 0 0 1.3cm 0, width=0.45\textwidth]{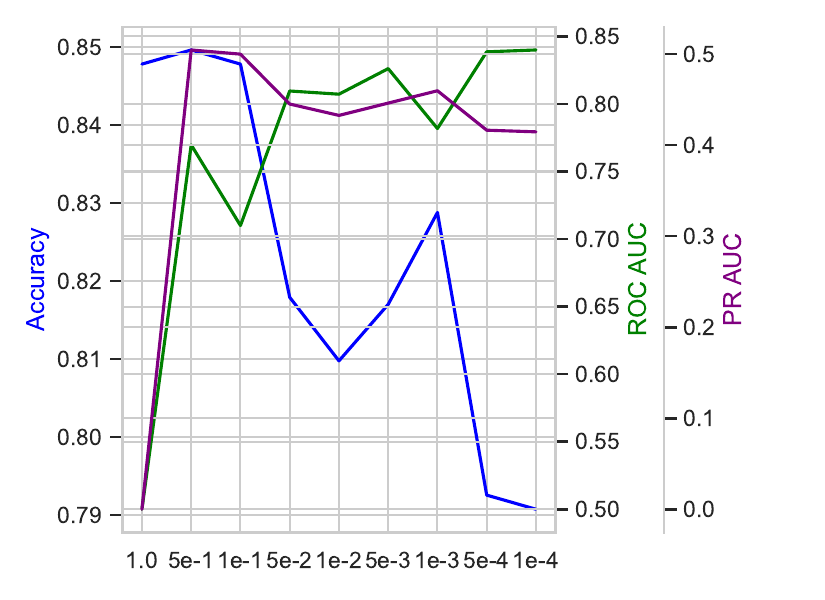}
    \caption{This figure shows an ablation study to find the optimal value of $\lambda$ decay used in the auxiliary loss.}
    \label{fig:decay}
\end{figure}

\subsubsection{Comparison with Advanced CapsNets}
In comparing CardioCaps with advanced CapsNet-based models, we have selected EM-CapsNets and Efficient-CapsNet as our baselines. First, EM-CapsNet represents an advanced approach of DR-CapsNet, incorporating matrix-wise capsules and an EM-routing algorithm within a deeper architecture. Second, Efficient-CapsNet replaces dynamic routing with self-attention routing, similar to our CardioCaps. As shown in Table \ref{tab:final_results_caps}, CardioCaps consistently outperforms these two advanced methods when applied to echocardiogram datasets. This highlights the efficacy of the novel methods introduced in CardioCaps, specifically the new loss function and attention mechanism, in effectively managing class-imbalanced echocardiogram datasets.

\subsubsection{Ablation Study}
In Section \ref{sec3}, the results of the ablation study demonstrate that CardioCaps performs optimally when all components, including the attention mechanism, the affine matrix, and the weighted margin loss function, are incorporated. The removal of any of these components results in decreased performance. In terms of the new loss function, CardioCaps shows improved performance with the weighted margin loss function in addition to L2 regularization loss. Regarding $\lambda$ decay in the regularization loss, as illustrated in Figure \ref{fig:decay}, the study indicates that the model performs best when $\lambda=0.5$ in the specified range [1e-4, 1.0).

\section{Conclusion}
In this paper, we propose \textit{CardioCaps}, a model designed for class-imbalanced echocardiogram classification. The network comprises two main components: a weighted margin loss with an L2 regularization loss function derived from the EF regression task and an attention mechanism as a substitute for dynamic routing. The new loss function addresses the scarcity issue of positive cases in class-imbalanced echocardiogram datasets by assigning them more significant importance. Adopting the attention mechanism enhances the efficiency of calculating the similarity between lower and upper capsules compared to dynamic routing. We evaluate CardioCaps using publicly available echocardiogram datasets and demonstrate its performance surpasses various state-of-the-art baseline methods. 

\clearpage
\section*{Acknowledgment}
This research was supported by the [Bio\&Medical Technology Development Program] of the National Research Foundation (NRF) funded by the Korean government (MSIT) (No. RS-2023-00222838).

\bibliographystyle{ieeetr}
\bibliography{main}

\end{document}